# Institutional Newspapers Pipeline: Deriving billions of high quality tokens from historical newspapers

**Matteo Cargnelutti** [a], **Catherine Brobston** [a], **Eben English** [b], **Jake Sadow** [b], **Kacie Bailey** [a], **Greg Leppert** ✉ [a], **Amanda Watson** [c], **Jessica Chapel** [b], **Jonathan Zittrain** [d]

[a] Institutional Data Initiative, Harvard Law School Library
[b] Boston Public Library
[c] Harvard Law School Library
[d] Harvard Law School, Harvard School of Engineering and Applied Sciences, Harvard Kennedy School

## Abstract

Historical newspapers are an abundant record of public life, but their dense, irregular and sometimes noisy layouts make computational access to these materials both challenging and limited. We present the Institutional Newspapers Pipeline, a modular system we jointly designed with Boston Public Library to extract high-quality, structured datasets from historical newspaper scans. It was architected so that each step remains interpretable and customizable, and so that the pipeline as a whole remains computationally frugal enough to run on workstation-level hardware. The pipeline runs each scan through a multi-step process: it segments scans into individual type-agnostic crops and performs OCR on each resulting segment before then performing text analysis, type classification, reading order detection, named entities recognition, subject classification, language detection, and pre-computed embeddings generation on every crop. We ran this pipeline against a portion of Boston Public Library's holdings and released the results as an open dataset. The optical character recognition (OCR) output represents 16.3 billion `o200k_base` tokens across 83.1 million individual crops, extracted from 1,473,635 public domain newspaper scans published between 1795 and 1930. This report describes our methods for each processing step, the small models we trained, as well as the evaluation results and dataset-scale measurements we collected in the process. It accompanies the release of the pipeline, models, and dataset. We position this work as a substantial step towards unlocking high-quality data from tens of millions of newspaper scans.

✉ Corresponding author: gleppert@law.harvard.edu

# 1 Introduction

It is notoriously hard to extract data from historical newspapers, especially at scale (Clausner et al., 2015; Nikolaidou et al., 2022). Their layouts are dense and irregular, source scans vary widely in quality, and the text is set in columns, decorative headings, and tables that general-purpose tools do not handle well. At the same time, libraries across the world dedicate significant effort to preserve, digitize, and provide access to historical newspaper collections. The Library of Congress' Chronicling America program[1] is a prominent example, as are Europeana Newspapers[2] and the Bibliothèque nationale de France's Gallica press portal[3] in Europe.

Large language models (LLMs) require vast amounts of data in order to learn efficiently (Kaplan et al., 2020; Hoffmann et al., 2022). The nature and quality of what these models learn during pre-training influences their performance and behavior (Longpre et al., 2024). We posit that data derived from high-quality newspapers can significantly contribute to improving AI's "digital diet"[4], and that this work can enable novel use cases beyond model training—from patron access to research. Open pre-training corpora such as The Pile (Gao et al., 2020) and Dolma (Soldaini et al., 2024) demonstrate the value of releasing diverse pre-training data and processing recipes to the research community. The Talkie-LM project (Levine et al., 2026), which trains LLMs on pre-1931 data to better understand the prediction capabilities of LLMs, is one example of the downstream impact such data can have. Talkie-LM was trained, in part, on Institutional Books: Harvard Library (Cargnelutti et al., 2025), which we released in 2025.

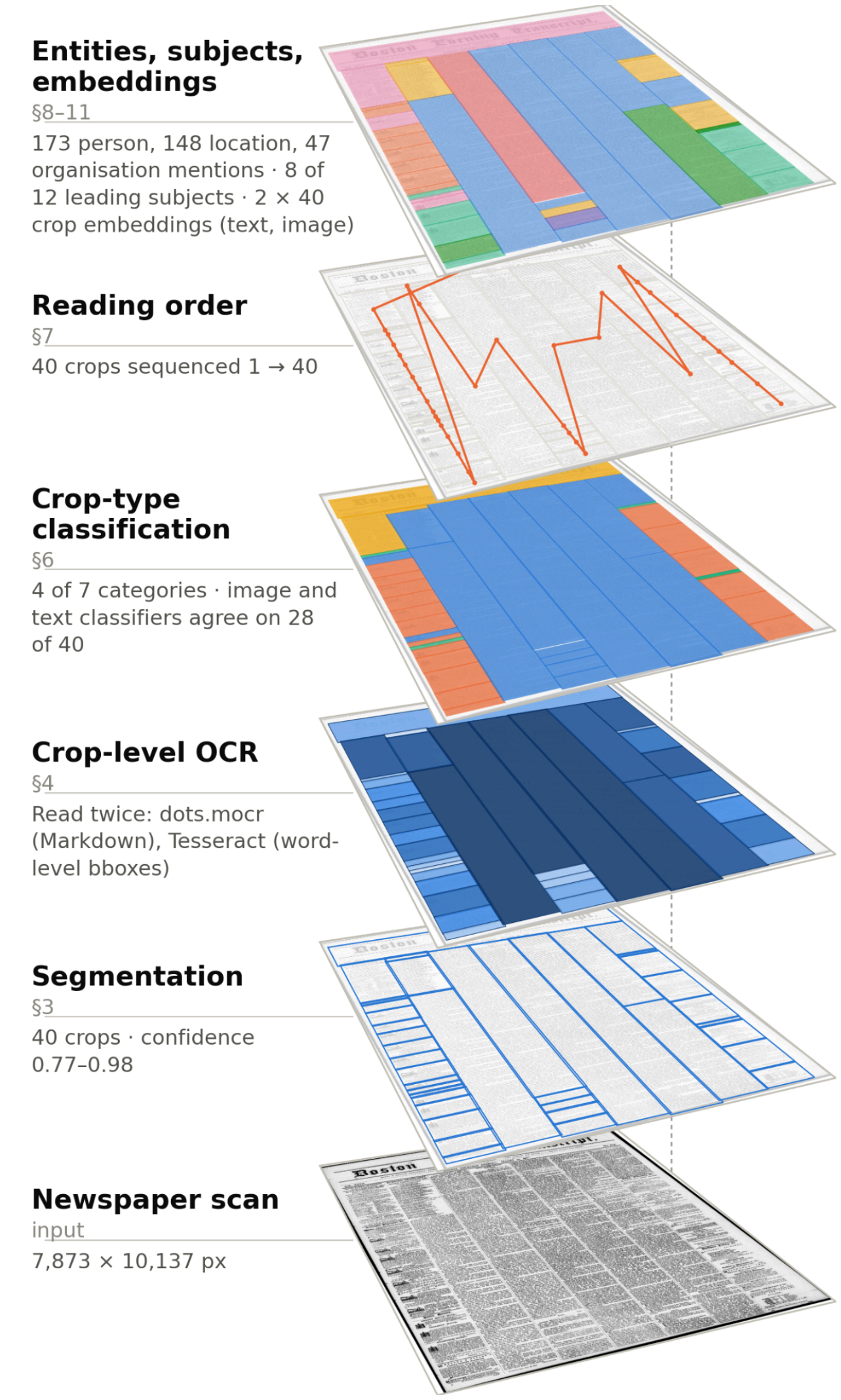

***Figure 1.*** *Output of the pipeline on the front page of the Boston Evening Transcript of 28 August 1856. The pipeline's fifteen steps are grouped here into six layers, shown from the bottom up: the source scan; the 40 crops that the segmenter detects; the OCR text of each crop, shaded by token count; the type of each crop; the reading order as a path through the crops; and the entities, subjects, and embeddings that the enrichment steps add.*

[1] https://www.loc.gov/collections/chronicling-america/
[2] https://www.europeana.eu/en/collections/topic/18-newspaper
[3] https://gallica.bnf.fr/selections/fr/html/presse-et-revues
[4] https://hls.harvard.edu/today/food-for-ai-thought-and-the-library-initiative-improving-ais-digital-diet/

Furthermore, research into the historical events recorded in these materials has been hindered by the relative inaccessibility of their contents (Nguyen et al., 2021; Ehrmann et al., 2023). For many newspaper collections, the extracted text is of such low quality that it serves primarily as a loose and unreliable keyword search used to locate the original page scans from which a human can then read, rather than as an accurate representation of the underlying source which can be mined and analyzed using computational tools (Traub et al., 2015). Better extraction of content from newspapers has the potential to unlock new insights into our past at a scale only recently made possible by these emerging computational tools, including LLMs. And where LLMs have been trained on high-quality representations of historical materials, they show increased capabilities for analyzing them (Manjavacas and Fonteyn, 2022).

This newspaper processing pipeline was jointly designed by the Institutional Data Initiative and Boston Public Library. Our collaboration was guided throughout by Boston Public Library's focus on improving patron and researcher access through seeking an “atomic” understanding of individual items within its newspaper scans collection.

In designing this pipeline, we took inspiration from the work conducted by Lee et al. (2020) on the Newspaper Navigator dataset and Dell et al. (2023) with American Stories. Both explored using convolutional neural networks (CNNs) (LeCun et al., 1998) to segment and classify newspaper layout items.

While prior art proved how effective CNNs can be in this context, we chose to approach the problem from a different angle, informed by the unique challenges of the collection at hand and Boston Public Library's goals. We defined an atomic newspaper item as a “crop,” a standalone piece of content segmented from a newspaper scan. We separated segmentation from classification, both to better leverage our finite volume of training data and to treat classification as a distinct problem. We combined text and image classifiers for crop-type classification to disambiguate hard-to-classify items where text or visual signal alone is not sufficient. More generally, we kept each step of the processing pipeline separate, so that both the Institutional Data Initiative and Boston Public Library teams could individually evaluate, interpret, and steer them, and so that each step remained customizable for other collections. Throughout, we balanced the need for high-quality data from complex objects against computational frugality (Vanderbauwhede, 2023), favoring small, efficient methods and models that keep the pipeline within reach of workstation-grade hardware.

This technical report describes the pipeline we created, the methods we developed, and the models we trained, as well as the dataset we generated out of 1,473,635 newspaper scans from Boston Public Library's collection. This report's figures focus on the pre-1931 subset of Boston Public Library's collection, which we released as part of our dataset. We position this work as an initial step towards unlocking high-quality data from tens of millions of newspaper scans. This release is the first iteration in this process.

Figure 1 shows a simplified view of what the pipeline produces on a single scan from the collection. Each layer holds the data that one group of steps adds to that same page. The sections that follow describe these steps in the order the pipeline runs them.

## 2 Contributions

With this technical report, we introduce this initial set of contributions:

- **A dataset** produced from our ongoing collaboration with Boston Public Library on its public domain newspapers collection. For each scan, we provide all of the data we collected and generated at crop level: bounding boxes, crop types, OCR outputs from both Tesseract and a specialized VLM, reading order indexes, NER entities, subject classes, as well as text and language analysis metrics. The VLM OCR output represents 16.3 billion `o200k_base` (OpenAI, 2024) text tokens for 83.1 million individual crops, extracted from 1,473,635 individual scans.
- **A production-ready pipeline** that knowledge institutions can use on their own newspaper collections.
- A series of small and efficient **detection and classification models** for processing large-scale historical newspaper collections. This report describes the training and evaluation processes of each model. Additional information can be found in each model's Hugging Face repository.

All of these contributions are available:

- On Hugging Face: https://huggingface.co/collections/institutional/institutional-newspapers
- On GitHub: https://github.com/institutional/institutional-newspapers-pipeline

An Agent Skill file[5] is also available to enable agentic use of this dataset and the accompanying models.

## 3 Segmentation

### 3.1 Methodology

As a first step towards reaching an "atomic" understanding of newspaper scans, we set a goal of segmenting pages into individual crops. In the context of this project, a crop is a rectangular bounding box around an uninterrupted flow of text or visual elements. As a result, headlines are not separated from the article they anchor, and visual-heavy advertisements are generally self-contained within a single crop.

Taking inspiration from prior art, we hypothesized that separating crop segmentation from classification could help us better leverage a limited number of annotations. The segmentation model needed to learn from—and run inference against—full scans. As content and advertisement crops represent the vast majority of crops on each scan (see Section 6), training a combined detection-plus-classification model on full scans would have led to significant class imbalance, likely degrading both detection and classification accuracy. This concern comes in addition to a classification challenge we had previously identified in which certain crops could only be reliably classified using a combination of text and image signal (see Section 6). Separately, but for similar reasons, we assessed that an object detection model would likely perform better at that task than a segmentation model. The precision of the masks object detection models try to generate can conflict with the dense, noisy nature of newspaper scans.

We chose `YOLO26x` (Jocher et al., 2026) as our base model. While it is the largest model of the YOLO26 series at 55.7M parameters, it remains a fairly small CNN suitable for inference at scale.

[5] https://agentskills.io/

In order to train and evaluate this model, we annotated 1,020 randomly sampled newspaper scans from Boston Public Library's collection using the VGG Image Annotator (VIA) (Dutta and Zisserman, 2019), for a total of 47,485 individual bounding box annotations. 210 of these scans were fully annotated manually, while 810 were pre-annotated using intermediary models trained on all available annotations. All machine-generated annotations were manually reviewed and edited. Of the 47,485 boxes in our annotations set, 9,028 were drawn manually and 38,457 were pre-annotated by our intermediary models. This process allowed us to iteratively estimate how many annotated scans we needed in order to train an effective object detection model, and to progressively focus our attention on patterns the model struggled with.

The final model was trained and evaluated on 1,020 scans, with 15% of them set aside for validation and another 15% for evaluation. To balance computational expense and accuracy, we chose to train our model with scans at a resolution of 960 pixels[6], after iterating over resolutions ranging from 320px to 1280px. At inference time, we discard detections with a confidence score below 0.6 and apply non-maximum suppression with an intersection-over-union (IoU) threshold of 0.15. The confidence threshold removes low-certainty detections, while the deliberately low IoU threshold suppresses heavily overlapping boxes.

| **Metric** | **Value** |
|---|---|
| Test scans | 153 |
| Test instances (boxes) | 2,991 |
| Precision | 0.927 |
| Recall | 0.910 |
| F1 | 0.918 |
| mAP@50 | 0.955 |
| mAP@50-95 | 0.901 |

***Table 1.*** *Segmentation model evaluation on the held-out test set.*

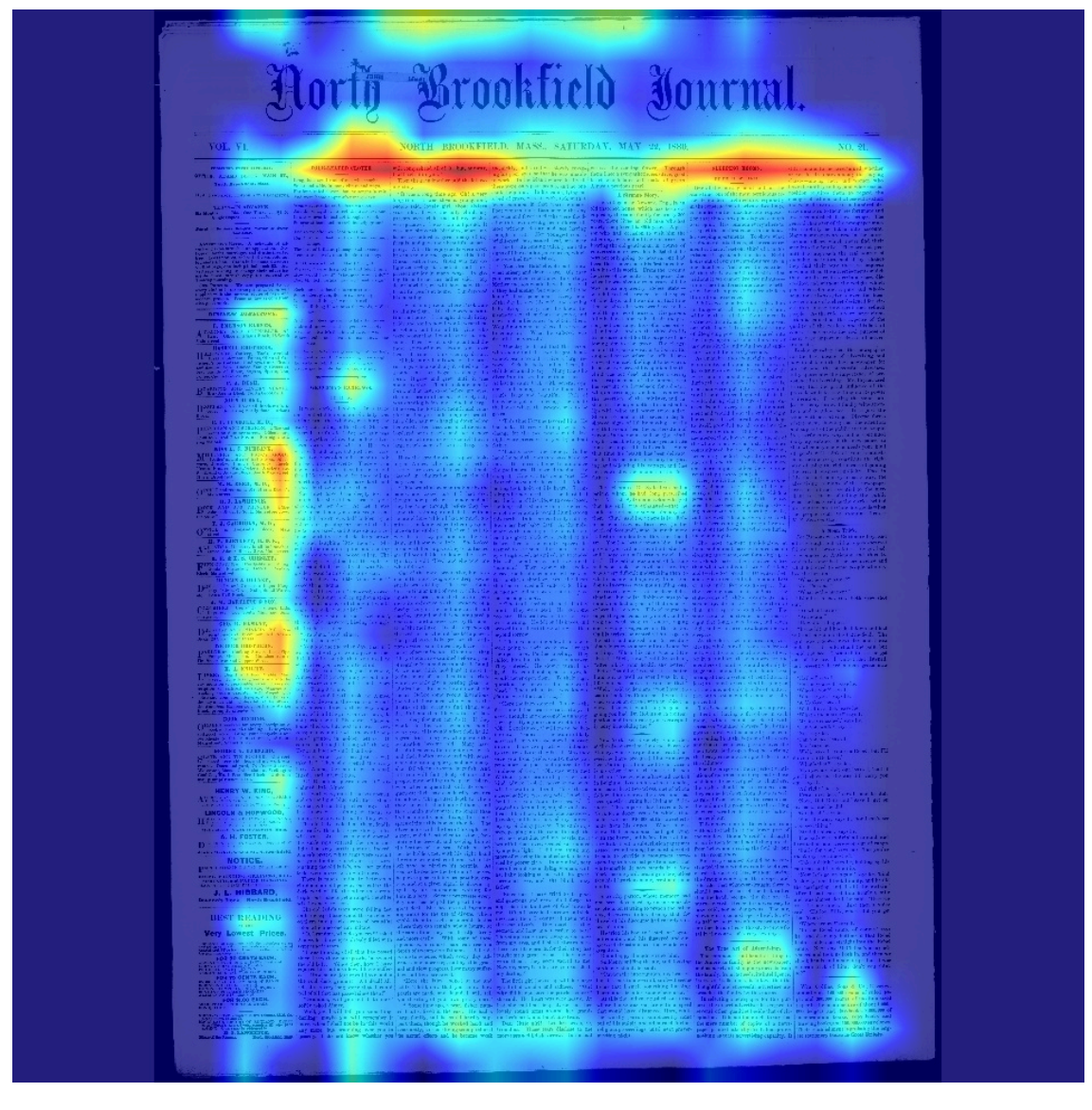
North Brookfield Journal.

***Figure 2.*** *Eigen-CAM heatmap for the segmentation model, overlaid on a sample front page (North Brookfield Journal, May 22 1880). Computed at backbone layer index 8 (a* `C3k2` *block) at an inference size of 960 pixels. Activation concentrates on the masthead and headline banner and traces the column structure of the page.*

[6] More information on this training run is available on Hugging Face: https://huggingface.co/institutional/institutional-newspapers-segmenter-yolo26x/blob/main/args.yaml

Our evaluation set of 153 held-out scans (2,991 bounding-box instances) suggests that the model effectively segments this collection. Table 1 summarizes these results.

To help interpret the model's behavior, we generated Eigen-CAM heatmaps (Muhammad and Yeasin, 2020; rigvedrs, 2026) from the trained segmenter. Eigen-CAM projects the activations of a chosen layer onto their first principal component which highlights the regions that contribute most to the model's response.

Figure 2 overlays such a heatmap on a sample front page, computed at index 8 of the model's backbone (out of 11), the deepest block whose receptive field is still purely convolutional. The strongest activation follows the masthead and the headline banner across the top of the page. Secondary activation traces the vertical column structure and the denser advertisement block in the left margin. This suggests that the model learned to respond to the structural cues of a newspaper page (the layout boundaries between columns and blocks) rather than to incidental features of the scan.

## 3.2 Results

Out of 1,473,635 scans, the model identified 83,147,041 individual crops, for an average of 56.4 crops per scan. Model confidence is highly concentrated at the top of the range, around 0.95 (Figure 3). This is consistent with the conservative confidence threshold applied at inference time.

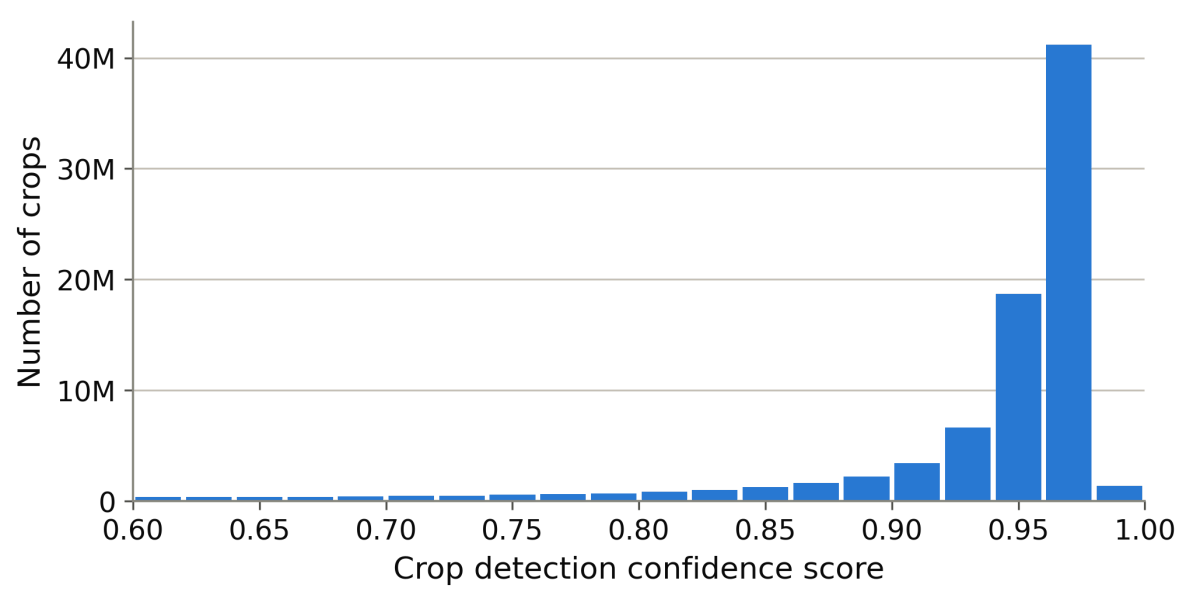

***Figure 3.*** *Distribution of crop detection confidence scores across all detected crops.*

***Figure 4.*** *Average number of crops per scan, by decade.*

In this collection, mid-19th-century newspaper scans appear to be the most dense in terms of average number of crops per scan. While it may be tempting to draw conclusions about the nature of these scans and the way newspaper layouts evolved, further analysis would be needed to clearly identify the cause (Figure 4). The data this pipeline produces may help digital humanities researchers investigate such questions.

Because our pipeline filters detections with low confidence scores and uses IoU to eliminate overlapping boxes, an average 89% of each scan is covered by crop detections (Figure 5). That figure suggests both that our model may have missed certain crops, and also that it did not generate crops for margins. Measuring the outer margins of a sample of scans supports the second explanation: the outermost crops leave an uncovered border of 2.6% to 3.6% of the scan on each side, which appears to account for most of the uncovered area (Appendix A).

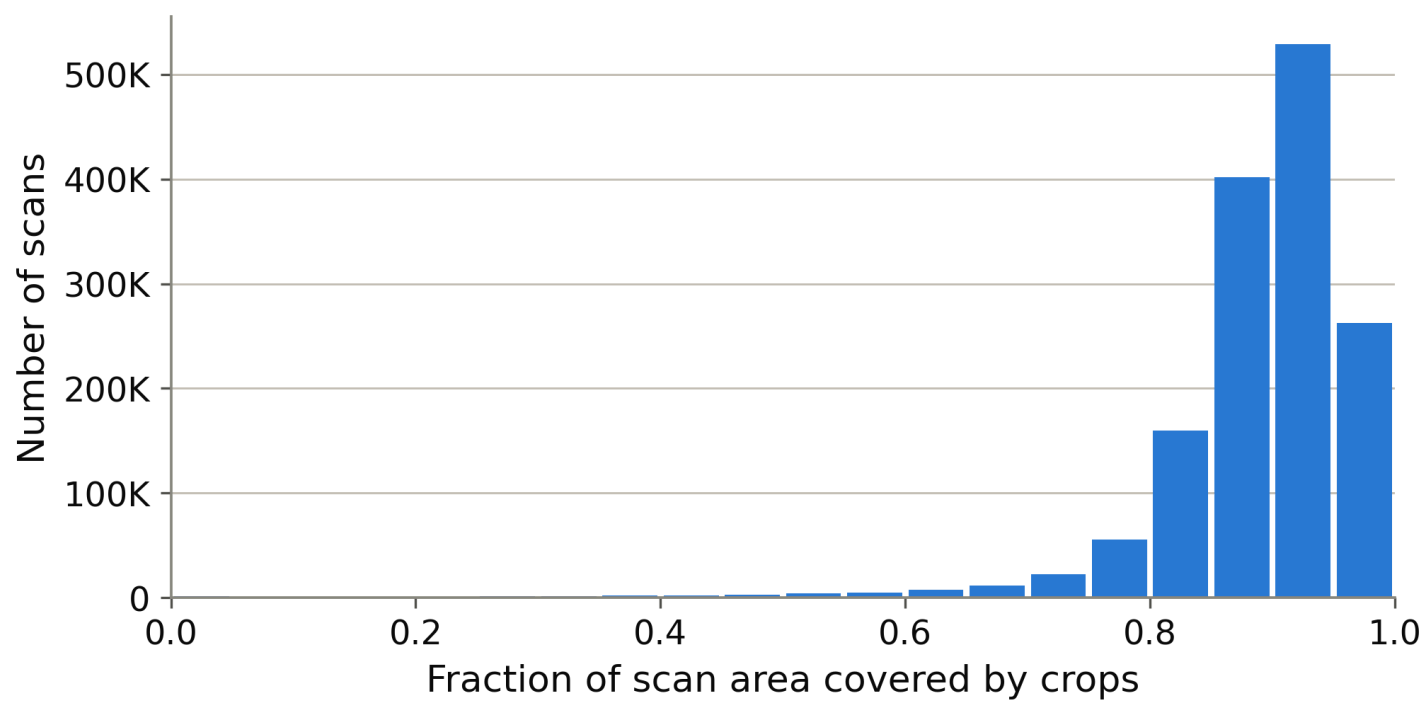

***Figure 5.*** *Distribution of the fraction of scan area covered by detected crops.*

# 4 Crop-level OCR

## 4.1 Methodology

OCR is one of the key challenges involved in extracting high-quality data from newspaper scans (Nguyen et al., 2021). During our initial tests, we observed that traditional OCR engines often failed to properly detect columns and parse large, decorative text elements. OCR-capable vision-language models (VLMs), which often produce higher-quality extractions than traditional open-source tools (Poznanski et al., 2025) yielded promising results out of the box when run against full scans, but accuracy and performance were inconsistent and seemed to drop significantly on complex or dense scans. This observation seems consistent with prior work. Layout complexity and line count are among the strongest predictors of VLM transcription error on historical pages, particularly for smaller and weaker models (Levchenko, 2025), page-level evaluation on historical corpora leaves even frontier models well short of production accuracy (Semnani et al., 2025), and on 19th-century newsprint, passing large text regions to a VLM without splitting them into smaller, rectangular crops induces degenerate repetition loops that push mean character error rates far above the median (Bourne, 2025). Reproducibility, operational control and cost were also a concern. Using frontier models, the cost of the 16 billion tokens we needed to generate would have ranged approximately from $250,000 to $800,000[7], assuming a cost per million output tokens ranging from $15 to $50. Instead, operating at the crop level with local models and techniques let us assemble a reproducible system we could deploy and run at a fraction of this projected cost (see Section 12.2), with granular control over the system's input, output and behavior.

We therefore chose to focus our efforts on crop-level OCR, and to do so with both Tesseract and a specialized VLM.

We chose to use Tesseract 5 (Smith, 2007; Tesseract OCR contributors, 2024a) and the `tessdata_best` models (Tesseract OCR contributors, 2024b). The failure modes of “traditional” OCR engines are well understood (Rice et al., 1999), and, contrary to VLMs, “hallucination” rarely occurs at the level of a full word or sentence (Yang et al., 2025). In addition, the word-level bounding boxes Tesseract produces can be used to compile OCR outputs compatible with existing library discovery systems (e.g, AltoXML[8]).

[7] Estimated at the time of writing this technical report (August 2026)

For the VLM portion of this apparatus, we tested multiple specialized VLMs for OCR and chose `dots.mocr` (Zheng et al., 2026) after reviewing samples. `dots.mocr` is only 3B parameters, offers multilingual support and, importantly, can parse tables. Our initial tests were run with `dots.ocr` (Li et al., 2025), which we replaced with `dots.mocr` shortly after it was released. This model is not immune to common pitfalls of VLM-based OCR including hallucinations (Ji et al., 2023) and token loops; in the latter, the model repeats the same token ad infinitum (Yang et al., 2025; Holtzman et al., 2020). It is worth noting that some of these "hallucinations" can be helpful, for example when transcribing damaged text (see Figure 6).

| **Tesseract** | **dots.ocr** |
| --- | --- |
| `TOR \GE Ste age Revs s may he found on - applicationat I %e Advertiser Office.` | `STORAGE Storage-Rooms may be found on application at The Advertiser Office.` |

***Figure 6.*** *Example of a helpful "hallucination", in which the VLM was able to recover missing letters from a damaged scan. The same mechanism can lead to transcription errors.*

We used the `tesserocr` Python library (sirfz, 2024) to run Tesseract against individual newspaper crops. This particular library allowed us to leverage multiprocessing to scale operations up to the entire collection. In order to balance accuracy with computational footprint, and after multiple rounds of quality testing, we chose to scale the image size of each crop down to a maximum of 1.5 megapixels before passing it to Tesseract.

Optimizing for inference with `dots.mocr` proved challenging and required making important decisions in advance. We ran inference using vLLM (Kwon et al., 2023) against 8 NVIDIA L40S GPUs, with each GPU running its own vLLM process.

We clamped image size for each crop to between 0.25 and 1 megapixel which appeared to offer an appropriate accuracy-versus-performance trade-off in that specific scenario. We also made aggressive use of vLLM's batching and caching capabilities and implemented a double-buffering mechanism in which the next batch is being prepared while vLLM runs inference on the current batch.

[8] https://www.loc.gov/standards/alto/

## 4.2 Results

We counted OCR tokens with the `o200k_base` tokenizer to give a model-relevant measure of text volume. Across the collection, `dots.mocr` produced 16.3 billion tokens and Tesseract 14.7 billion. With few exceptions, `dots.mocr` consistently produced more tokens than Tesseract, both in total (Figure 7) and per scan (Figure 8), which is consistent with its ability to recover text from dense crops, decorative elements, and damaged scans that Tesseract tends to struggle with.

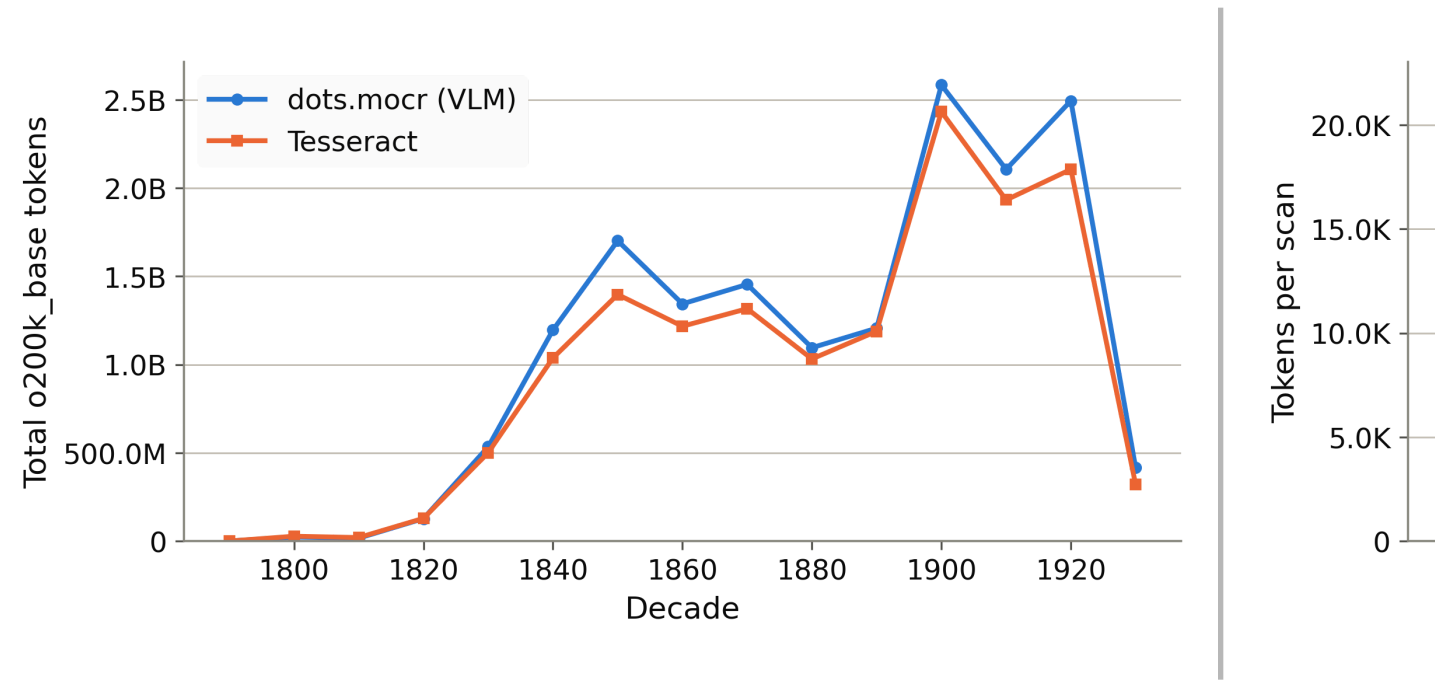

*Figure 7. Total `o200k_base` tokens over time (by decade), comparing `Tesseract` and `dots.mocr`.*

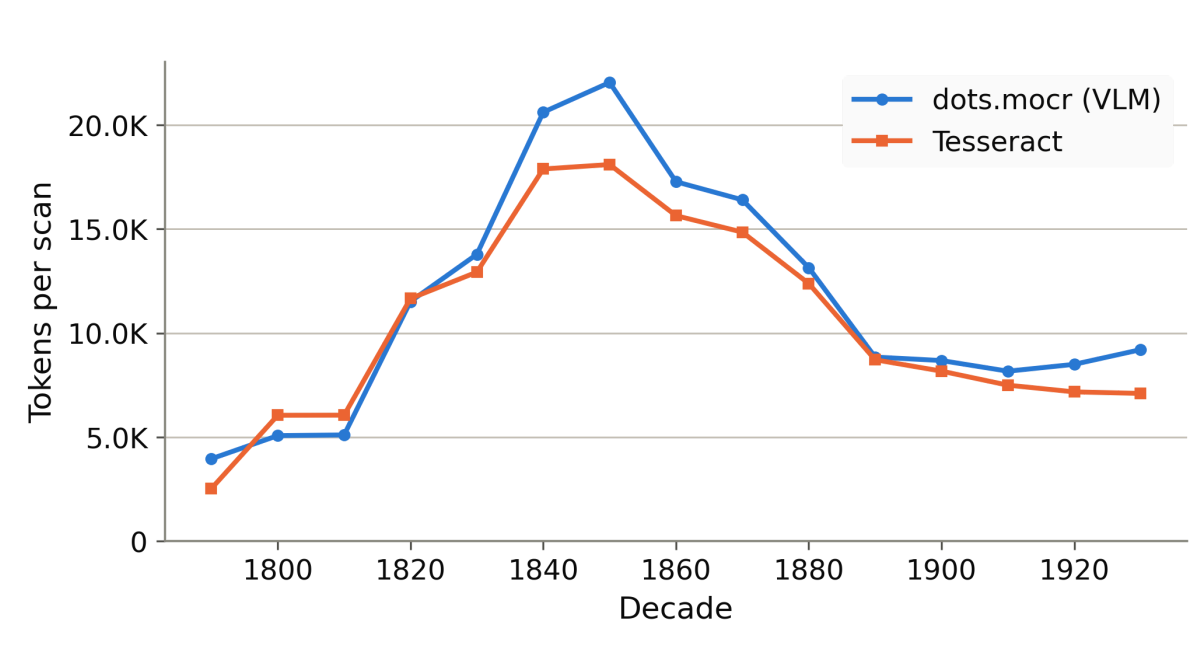

***Figure 8.*** *`o200k_base` tokens per scan over time (by decade), comparing Tesseract and `dots.mocr`.*

# 5 Crop-level Language Detection and Text Analysis

## 5.1 Methodology

### 5.1.1 Text analysis

In order to measure and compare the performance of the two OCR techniques we used, we collected the following metrics for the OCR text of every crop, for both Tesseract and `dots.mocr`:

- Character count.
- Word and sentence count, both total and unique, split with `PyICU` (PyICU contributors, 2024).
- A "tokenizability" score as a measure of how efficiently the extracted text is represented by the tokenizer.
- For `dots.mocr` output only: whether the text contains Markdown markers and/or a table.

We "flattened" the text of each crop before measuring it, so that the two OCR outputs could be compared on equal terms. Flattening strips HTML markup, rejoins words that were split by a hyphen at the end of a line, and replaces line breaks with spaces. We then removed zero-width spaces and split the result into words and sentences using `PyICU`, with the detected language of the crop as a processing hint a fallback to English when no language was available. The "tokenizability" score compares the number of words to the number of tokens those same words produce, on the basis that one word tokenizes into roughly 1.25 tokens for English text. This score can be used as a very coarse measure of text validity, though the presence of unusual vocabulary on which the tokenizer might not have been trained may negatively impact the score. Finally, we detect Markdown markers and tables on the raw VLM output rather than the flattened text, because flattening removes the very markers for which we look.

These metrics were also designed to offer continuity with the Institutional Books pipeline, with the goal of helping users make informed decisions when using the data.

### 5.1.2 Crop-level Language Detection

We chose to use Lingua (Stahl, 2024) to perform language detection at crop level, in order to help identify multilingual newspapers.

We made this choice based on the following criteria:

- Newspaper-derived text varies in length widely and Lingua is designed to remain accurate on short text.
- We already had access to issue-level language metadata derived from Chronicling America (Library of Congress and National Endowment for the Humanities, 2007) for this collection. We could therefore substitute issue-level language metadata for low-confidence detections or detections performed on very short text (see Section 13).
- We wanted to limit computational expense for a data point for which we already had an issue-level reference, which is also why we chose to run language detection only on VLM-extracted text.

## 5.2 Results

A language code was assigned to 99.81% of all crops (82,988,421 crops). Lingua detected 96.96% of these codes directly. The remaining 3.04% were inherited from issue-level metadata, under the post-processing rules described in Section 13. English is by far the most prevalent language of the collection, accounting for 97.61% of crops with a language code.

The pipeline surfaces a meaningful multilingual tail led by Yiddish, German, Swedish, and French, reflecting the immigrant press held in Boston Public Library's collection (Table 2). The collection covers 73 distinct language codes in total, though the distribution drops off sharply: the ten most common account for 99.97% of crops with a code, and the remaining 63 for 0.03% between them.

| Language | Crops | % of crops with a language code |
|---|---|---|
| English (ENG) | 81,002,114 | 97.61% |
| Yiddish (YID) | 983,375 | 1.18% |
| German (DEU) | 467,929 | 0.56% |
| Swedish (SWE) | 380,078 | 0.46% |
| French (FRA) | 76,217 | 0.09% |

***Table 2.*** *Top five language codes for `dots.mocr` (VLM) OCR text, as a share of crops with a language code.*

Comparing the two OCR methods on the same crops (Table 3), `dots.mocr` recovers more characters, words, and sentences than Tesseract and reaches a slightly higher mean tokenizability score. Tesseract reports more unique words, which is consistent with its higher rate of character-level noise producing spurious tokens.

`dots.mocr` also emits a measurable amount of structured output: 0.78% of crops contain a Markdown table and 3.33% contain other Markdown markup. Based on our observations, both of these metrics suggest that `dots.mocr` and our post-processing (Section 13) failed to format a non-trivial but hard-to-measure amount of tables and headings.

| Metric | Tesseract | dots.mocr (VLM) |
|---|---|---|
| o200k_base tokens | 14,661,019,181 | 16,302,004,429 |
| Characters | 50,124,785,558 | 57,428,175,900 |
| Mean words per crop | 106.4 | 115.3 |
| Mean unique words per crop | 68.8 | 64.3 |
| Mean sentences per crop | 7.7 | 9.6 |
| Mean unique sentences per crop | 7.5 | 8.6 |
| Mean “tokenizability” score | 86.85 | 87.35 |
| Crops with a Markdown table | N/A | 647,301 (0.78%) |
| Crops with Markdown markup | N/A | 2,772,880 (3.33%) |

***Table 3.*** *Text-metrics comparison between Tesseract and dots.mocr across all crops. Means are measured over the crops that carried text for that engine, 81,133,206 for Tesseract and 82,985,118 for dots.mocr, out of 83,147,041 crops.*

# 6 Crop-type Classification

## 6.1 Methodology

Because we separated segmentation from classification, we were able to devise a method of crop classification that leverages both image and text signals at crop-level in order to help make a more accurate and interpretable prediction.

We therefore chose to train two small classifiers:

- An image classifier based on `YOLO26m-cls` (11.6M parameters).
- A static embedding model fine-tuned as a classifier. Specifically, we used `Model2Vec` (Tulkens and van Dongen, 2024) to fine-tune `potion-base-32M` (MinishLab, 2024).

For both models, we focused on broad categories (for example “Content”, “Advertisement”, “Photograph or illustration”) to limit the effect of class imbalance.

To prepare the training set, we used the FP8 variant of `Qwen/Qwen3-VL-30B-A3B-Thinking` (Qwen Team, 2025b), a vision-language model from the Qwen3 series (Qwen Team, 2025a), to auto-annotate a large set of randomly sampled crops. We manually reviewed sampled annotations to refine our auto-annotation prompt (Appendix B) and to confirm that classifiers could be trained on this data. Our final manual check was performed on 600 annotations, yielding a 91% accuracy score for strict checks (fully correct only) and a 94.50% accuracy score for standard checks (fully correct plus ambiguous results combined).

We assembled training sets for our classification models by randomly sampling rows from our collection of auto-annotated crops. In total, our image classifier was trained on 188,477 annotations, and our text classifier on 185,900 annotations, both with a 70/15/15 split as a target. Although the annotations were pulled from the same pool, the text classifier received 2,577 fewer annotations than the image classifier. 1,740 of these were crops without text (automatically marked as "Empty" for text classification purposes), while the remaining 837 were "Empty" crops we manually augmented to train the image classifier by rotating them at 90°, −90° and 180° to help reduce the effect of extreme class imbalance for this category.

The image classifier was trained on crop images resized to 768 pixels, while the text classifier was trained on VLM-extracted OCR text (`dots.ocr`)[9]. The image classifier was trained on seven distinct classes, including an "Empty" class, while the text classifier was trained on six. The text classifier has no "Empty" class because a crop without OCR-able text is treated as "Empty" by convention rather than predicted as such. Table 4 compares the per-class evaluation results of the two classifiers.

The models have significant overlap and clear differences. They agree closely on text-heavy categories such as Advertisement, Content, and Section heading. The image classifier is far stronger on visual categories. For "Photograph or illustration" it reaches an F1 of 0.84 against the text classifier's 0.48, and for "Cartoon" 0.92 against 0.68. This behavior matches our initial expectations since these categories carry little OCR-able text.

[9] More information about both training runs is available on Hugging Face:
https://huggingface.co/institutional/institutional-newspapers-crop-classifier-image-yolo26m-cls/blob/main/args.yaml
https://huggingface.co/institutional/institutional-newspapers-crop-classifier-text-model2vec/tree/main

| Category | Image P | Image R | Image F1 | Text P | Text R | Text F1 |
|---|---|---|---|---|---|---|
| Advertisement | 0.91 | 0.93 | 0.92 | 0.95 | 0.94 | 0.95 |
| Content | 0.93 | 0.90 | 0.92 | 0.90 | 0.96 | 0.93 |
| Section heading | 0.87 | 0.97 | 0.92 | 0.94 | 0.93 | 0.94 |
| Masthead, nameplate or running head | 0.98 | 0.79 | 0.87 | 0.93 | 0.86 | 0.89 |
| Photograph or illustration | 0.83 | 0.84 | 0.84 | 0.63 | 0.38 | 0.48 |
| Cartoon | 0.90 | 0.94 | 0.92 | 0.85 | 0.57 | 0.68 |
| Empty | 0.72 | 0.95 | 0.82 | N/A | N/A | N/A |
| Overall accuracy | 0.913 (Top-1) | | | 0.92 | | |

***Table 4.*** *Per-class evaluation of the image and text crop-type classifiers.*
*"Empty" is marked N/A for the text classifier, which does not model that class.*

In order to decide on the most likely category for any given crop, we combined both signals. The filtering mechanism we implemented makes decisions as follows:

- By default, priority is given to the classification with the highest confidence score.
- If the image classifier returns "Photograph or illustration", priority is given to the image classifier.
- If the image classifier returns "Empty", priority is given to the image classifier.
- If a crop's text is flagged as "Empty", priority is given to the image classifier.

We provide raw data from both the image and text classifiers as part of our dataset, so end users can apply a different filtering mechanism.

## 6.2 Results

Across the 83,147,041 crops we classified, the two classifiers agree on 90.55% of crops (Table 5). The combined decision follows the image classifier on 95.03% of crops and the text classifier on 95.51%. The two classifiers overlap heavily but neither reproduces the final decision alone, which confirms the relevance of combining both signals.

Both classifiers report high confidence on average: 0.953 for the image classifier and 0.967 for the text classifier, weighted across all crops. Confidence varies by category and is markedly lower for the visual categories both models handle least well. Appendix C reports the per-category figures.

| Measure | Crops | % of classified |
|---|---|---|
| Total classified crops | 83,147,041 | N/A |
| Image and text agree | 75,287,373 | 90.55% |
| Final category = image classifier | 79,018,344 | 95.03% |
| Final category = text classifier | 79,416,070 | 95.51% |

***Table 5.*** *Crop-type classification agreement and the source of the final decision.*

By final category, advertisements are by far the most common crop type, followed by content (Figure 9). Conversely, content crops carry the majority of the OCR tokens, well ahead of advertisements (Figure 10). This pattern holds over time. Although advertisement crops outnumber content crops in most decades, content still accounts for the majority of tokens in nearly all of them (Figures 11 and 12).

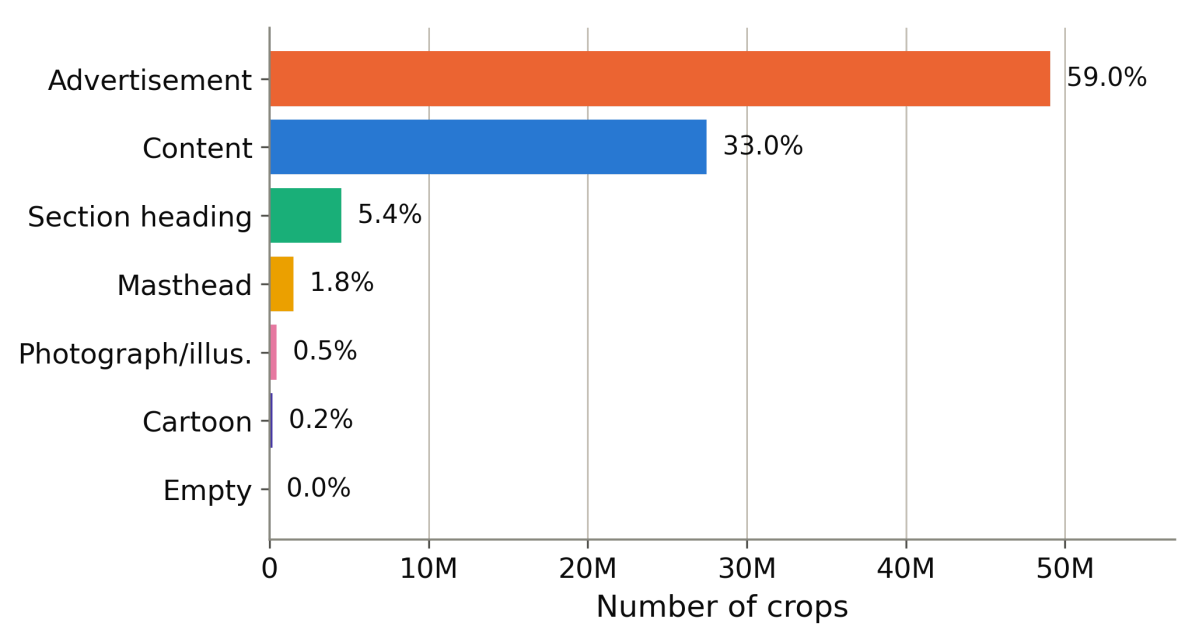

***Figure 9***. *Distribution of crops by final crop-type category.*

***Figure 10.*** *Total dots.mocr o200k_base tokens by final crop-type category.*

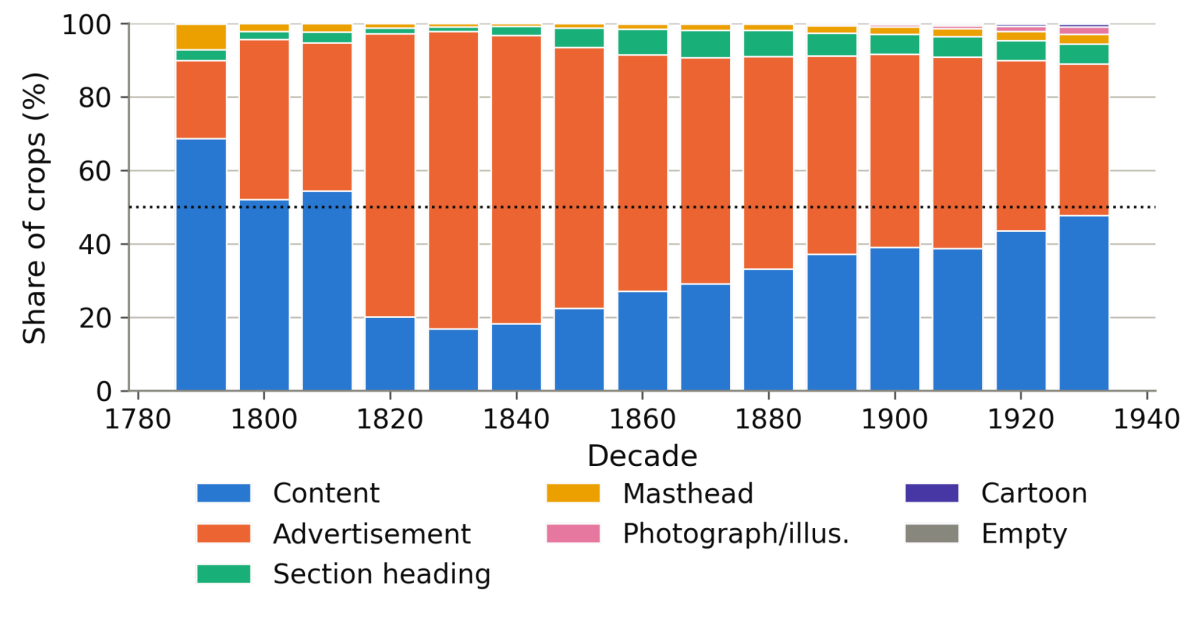

***Figure 11***. *Share of crops by final category over time (relative %, by decade).*

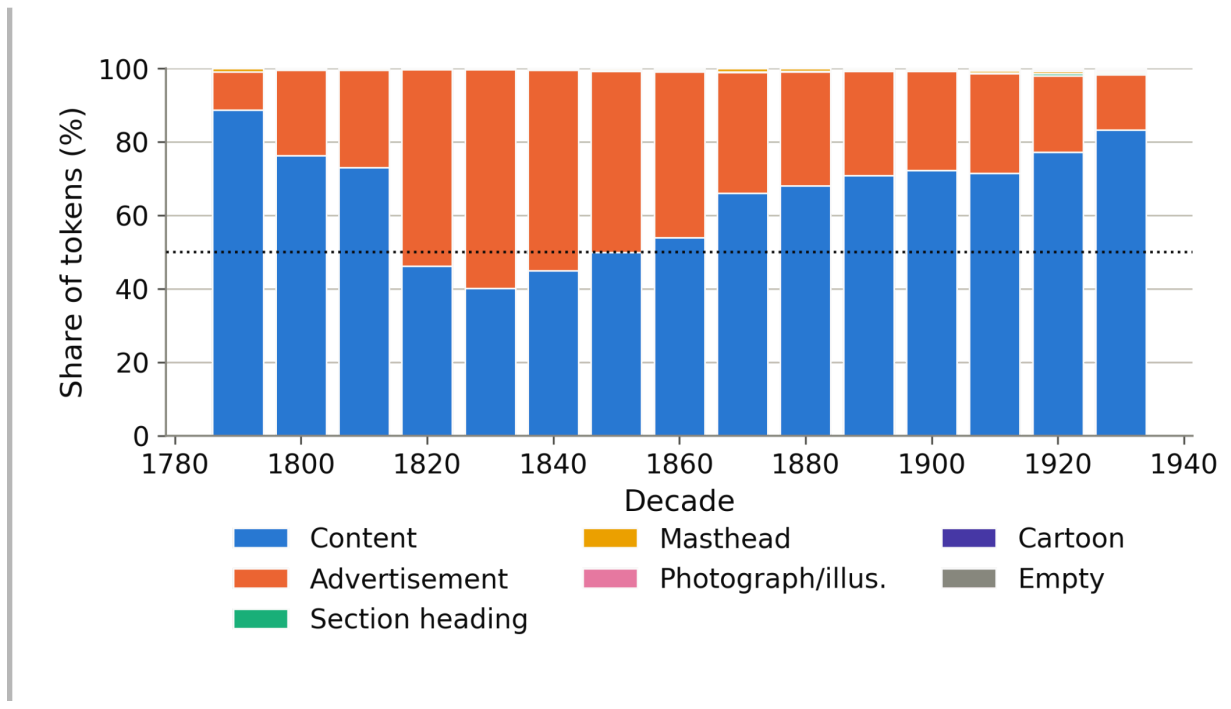

***Figure 12.*** *Share of dots.mocr tokens by final category over time (relative %, by decade).*

Looking across decades at the total scan area each category covers (Figure 13) shows how the relationship between crop surface area and token density evolves in this collection. Until the 1860s, advertisements took up roughly the same share of the page as they contributed in text tokens. From the 1870s onward, advertisements hold meaningfully more surface area than they contribute in text tokens, and the gap widens to about 15 percentage points by the 1920s. This is consistent with advertisements becoming progressively more visual over the period.

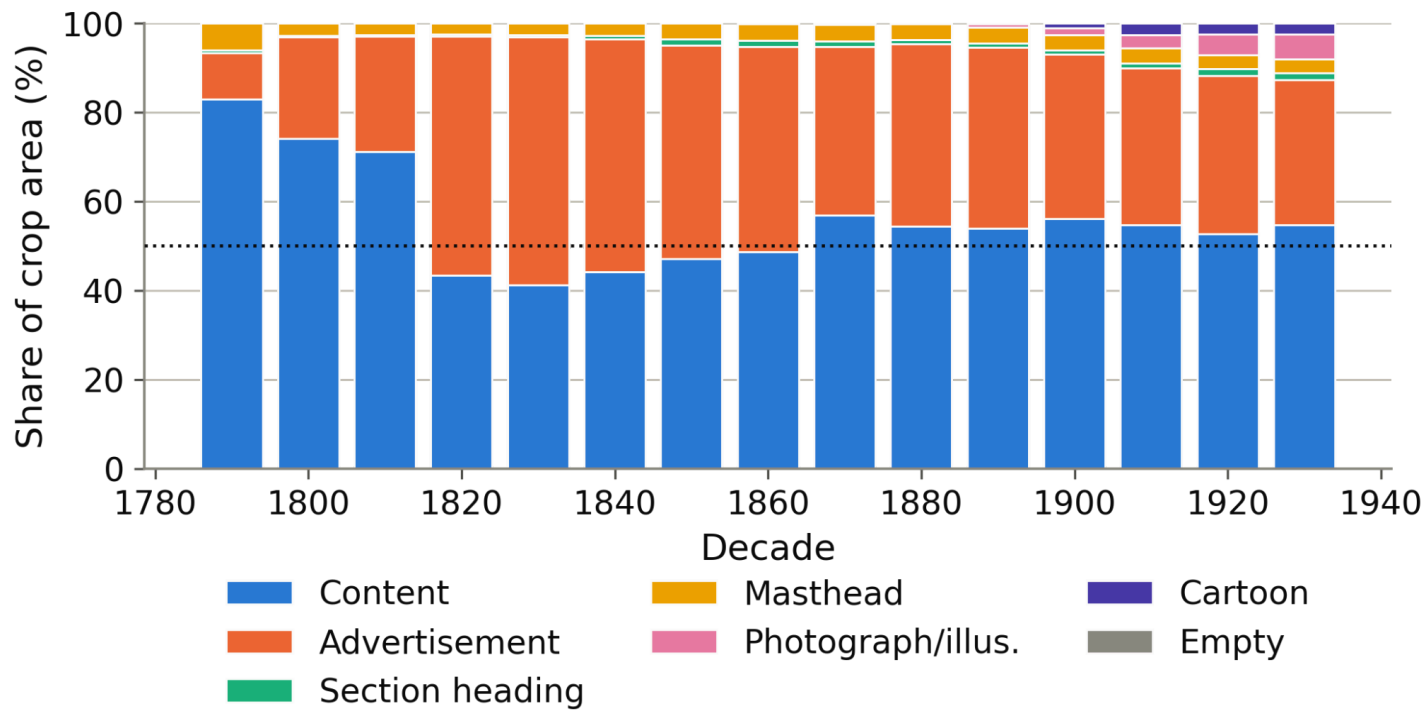

***Figure 13.*** *Share of crop area by final category over time (relative %, by decade).*

# 7 Scan-level Reading Order Detection

## 7.1 Methodology

Inferring the natural reading order of layout elements is known to be challenging (Wang et al., 2021). Our newspapers processing pipeline works at the level of crops. On any given page, only a subset of crops are directly connected to each other (for example, different parts of an article spread across different columns or pages), which arguably makes this problem more difficult to solve.

Newspaper layouts vary widely, not only over time but also from one title to the next. We identified early in our prototyping that applying a purely rule-based reading order detection method would not yield satisfactory results. Projects such as `LayoutReader` (Wang et al., 2021) explored using transformer-based models to perform reading order detection. While that approach yielded promising results, it presented two main issues for our use case. The first was that using a transformer-based model for this task at the scale of a library collection represents a significant computational expense. The second was that `LayoutReader` and similar models tend to operate at the level of spans, not crops.

We implemented a fast reading-order detection mechanism based on HDBSCAN (Campello et al., 2013; McInnes et al., 2017) combined with content-aware post-processing:

1. Classify crops as narrow or wide based on estimated column width relative to scan dimensions.
2. Cluster narrow “Content” crops into columns (HDBSCAN on x-centers).
3. Assign narrow non-content crops to the nearest column (by x-center distance).
4. Assign wide crops to the overlapping column with the most content above.
5. Order left-to-right by column, then top-to-bottom within columns.

6. Post-process via left-edge bucketing for final left-to-right, top-to-bottom “nudging”. Visual elements (photographs, cartoons) are sorted by their top edge rather than their y-center, so they appear before the content they illustrate.

To measure the accuracy of this reading-order detection method at page-level, we annotated 723 scans to establish an evaluation set. These reference scans were pre-annotated with an earlier version of this algorithm and corrected manually.

## 7.2 Results

Position accuracy is the share of crops the method places at exactly the right index in the sequence. We report it in two ways (Table 6). The macro figure scores each scan separately and then averages the scores without weighting them. The micro figure pools all crops and scores them together.

Measured against our evaluation set, the method reaches 72.1% macro and 80.8% micro position accuracy, with a Kendall's τ of 0.922 and 0.959 respectively.

| Crop type | Crops | Position accuracy (macro) | Position accuracy (micro) | Kendall's τ (macro) | Kendall's τ (micro) |
|---|---|---|---|---|---|
| Content | 14,859 | 0.716 | 0.744 | 0.970 | 0.974 |
| Advertisement | 21,630 | 0.671 | 0.849 | 0.820 | 0.955 |
| Section heading | 1,998 | 0.811 | 0.849 | 0.975 | 0.985 |
| Masthead, nameplate or running head | 872 | 0.935 | 0.917 | 0.932 | 0.938 |
| Photograph or illustration | 142 | 0.494 | 0.486 | 0.769 | 0.803 |
| Cartoon | 117 | 0.558 | 0.650 | 0.984 | 0.977 |
| Unknown | 939 | 0.629 | 0.765 | 0.853 | 0.936 |
| Empty | 12 | 0.667 | 0.667 | N/A | N/A |
| Overall | 40,569 | 0.721 | 0.808 | 0.922 | 0.959 |

***Table 6.*** *Reading-order evaluation: position accuracy and Kendall's τ, overall and per crop type. Macro figures average per-scan scores; micro figures are crop-weighted across all crops, across scans.*

We hypothesize that, in this context, the Kendall's τ metric is meaningful. While position accuracy assesses whether the predicted reading order matches our annotations exactly, the Kendall's τ figures suggest that the overall sequencing matches our annotations, even when exact positions differ.

While this method is overall effective and has a limited computational footprint, it cannot, by nature, properly handle deeply complex or unique layouts. We posit that these layouts would benefit from a transformer-based approach, and we envision this as a future direction for this project.

# 8 Crop-level Named Entity Recognition

## 8.1 Methodology

We chose to add a named entity recognition (NER) step to our pipeline in keeping with Boston Public Library's goal of achieving an "atomic" understanding of newspaper scans, as granular NER data can help enable new use cases for newspapers data (Ehrmann et al., 2023).

We consider this output to be experimental. Developing a NER method specific to this collection was out of scope for this initial pipeline development project and we therefore chose to use an "off-the-shelf" solution. We ran `Flair`'s base `ner-fast model` (Akbik et al., 2019) against the VLM-extracted text of each crop.

We applied the following filtering rules, designed through iterative testing:

- We discarded detections with a confidence score below 0.85.
- We only kept results for `PER` (persons), `LOC` (locations), and `ORG` (organizations) entities.
- We deduplicated detections and, for each group of likely duplicates, kept the detection with the highest confidence score.

## 8.2 Results

Across the collection, the pipeline detected 155.6 million location mentions, 142.2 million person mentions, and 49.1 million organization mentions (Table 7). Locations have the fewest unique surface forms relative to their volume, which is consistent with a small set of place names recurring across the collection.

The most frequent entities (Table 8) appear coherent with the nature of the collection: the locations are dominated by Boston and the surrounding region ("Boston", "Mass") alongside the major centers of national news ("New York", "Washington", "United States"), the organizations are dominated by institutions of government ("Congress", "Senate", "House"), and the persons are common family names of the period.

| Entity type | Total detections | Unique surface forms |
|---|---|---|
| PER | 142,245,435 | 381,205 |
| LOC | 155,570,972 | 298,058 |
| ORG | 49,059,445 | 380,115 |

***Table** 7. Named-entity totals and unique surface forms.*

| Rank | PER | LOC | ORG |
|---|---|---|---|
| 1 | Smith (612,147) | Boston (8,661,331) | Congress (600,270) |
| 2 | Brown (478,864) | New York (4,159,692) | WM (455,265) |
| 3 | God (389,782) | Washington (2,966,122) | Senate (427,325) |
| 4 | Clinton (314,286) | Mass (2,712,257) | House (385,912) |
| 5 | Johnson (308,811) | United States (1,888,415) | Union (365,444) |

***Table 8.** Top five entities by frequency, per entity type.*

This NER technique has limitations. Names may be out of distribution, in particular in historical contexts. OCR quality is another bottleneck. The model may also identify derogatory terms or other harmful language as names on occasion. For these reasons, while this data point can be used to assist with downstream research, it is experimental and should not be taken at face value.

# 9 Crop-level Subject Detection

## 9.1 Methodology

We decided to undertake crop-level subject detection as a way to provide additional signal on the nature of each crop. We consider this output experimental, and we focused mainly on establishing a process for performing such classification at this scale. The list of subjects we picked is based on our observations for this specific collection, and we believe it needs to be adjusted for specific research needs. We designed this section of the pipeline so it can be re-run as needed with a different set of subjects.

We ran `MoritzLaurer/ModernBERT-large-zeroshot-v2.0` (Laurer, 2024; Laurer et al., 2023), a fine-tune of `ModernBERT` (Warner et al., 2025), using Hugging Face's zero-shot classification pipeline (Wolf et al., 2020) on the VLM-extracted text of each crop, when available. We store the full “ranking” produced by the model, along with confidence scores.

## 9.2 Results

Across all crops, the most frequently top-ranked subjects are “Business, Finance & Market Reports” and “Commercial Advertisements & Classifieds”, followed by “Science, Technology, Health & Agriculture” and “Politics, International Relations & Government” (Figure 14). Two of these labels, “Commercial Advertisements & Classifieds” and “Mastheads, Page Headers & Printer Imprints”, overlap by design with the crop-type categories of Section 6. They were designed to give a secondary and independent signal on crop-type classification as well as to give the zero-shot classifier a broad category to fall back onto for ambiguous crops, rather than forcing a topical label onto a crop that carries no topic.

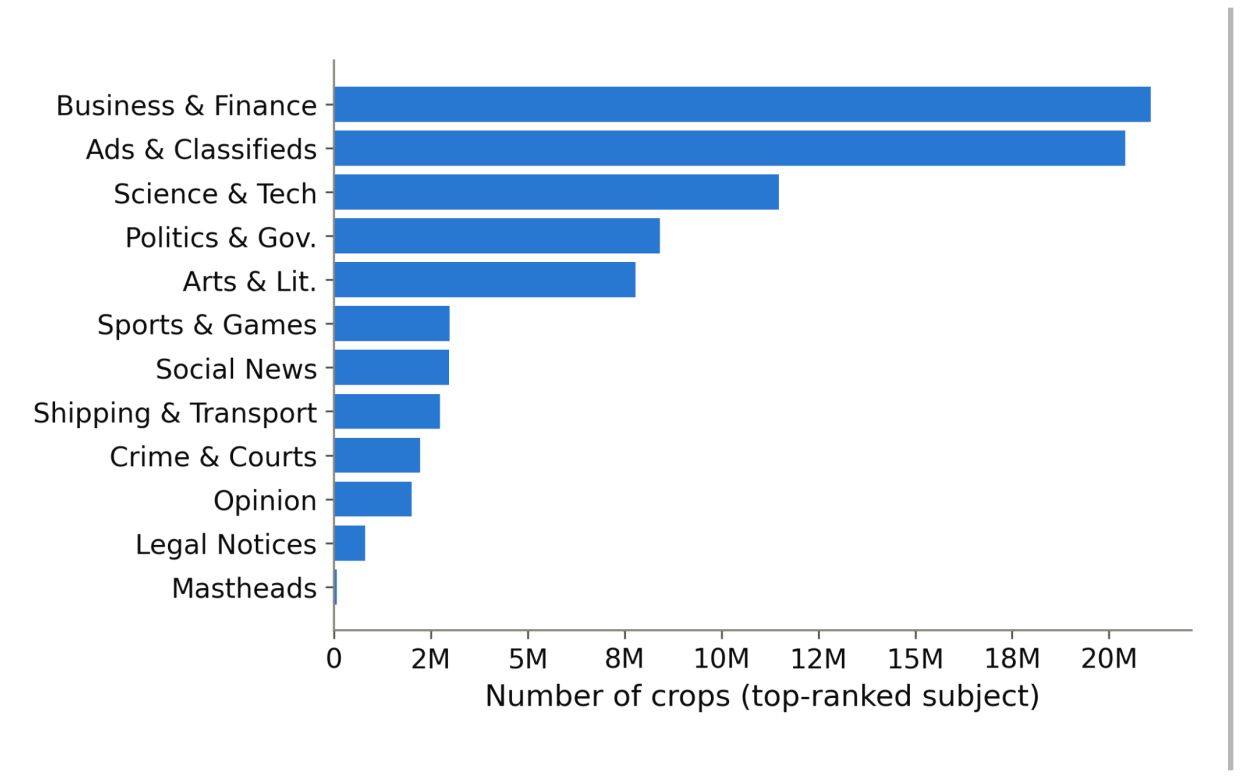

***Figure 14.*** *Top-ranked subject labels across all crops.*

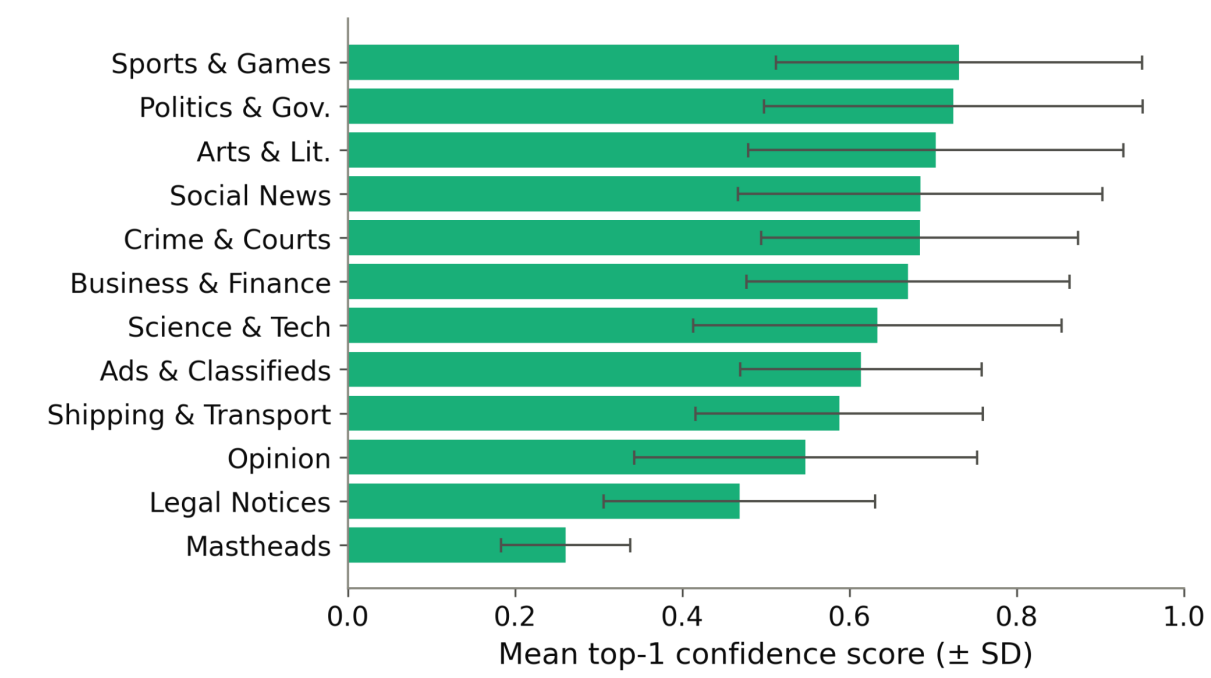

***Figure 15****. Mean top-1 confidence score per subject label, with standard-deviation error bars.*

Confidence scores are spread very widely within each subject (Figure 15). Across the 82,985,966 crops with a subject prediction, the mean top-1 confidence score is 0.65, but the standard deviation per label ranges from 0.08 to 0.23. Such a spread is expected in a zero-shot setting, where the classifier scores each label through textual entailment rather than against a distribution it was trained on (Yin et al., 2019). This is the main reason we treat this output as experimental. Mean confidence also varies by subject: the model is most confident on well-delimited topics such as “Sports, Racing & Games” and “Politics, International Relations & Government”, and least confident on broad or residual categories such as “Legal Notices, Auctions & Public Announcements” and mastheads.

The separation between the model's first and second choice is pronounced. Across crops with a subject prediction, the mean gap between the top-1 and top-2 confidence scores is 0.457, putting the mean second choice at 0.20. The runner-up is therefore well above chance, 1/12 across 12 labels. Nonetheless, because the spread is wide, the top labels are not equally reliable across crops, and users should work from the full distribution of scores. The dominant subject per crop type (Figure 16) behaves as expected: advertisement crops map overwhelmingly to commercial subjects, while content crops spread across news-oriented subjects, suggesting that the subject signal is overall consistent with the crop-type classification.

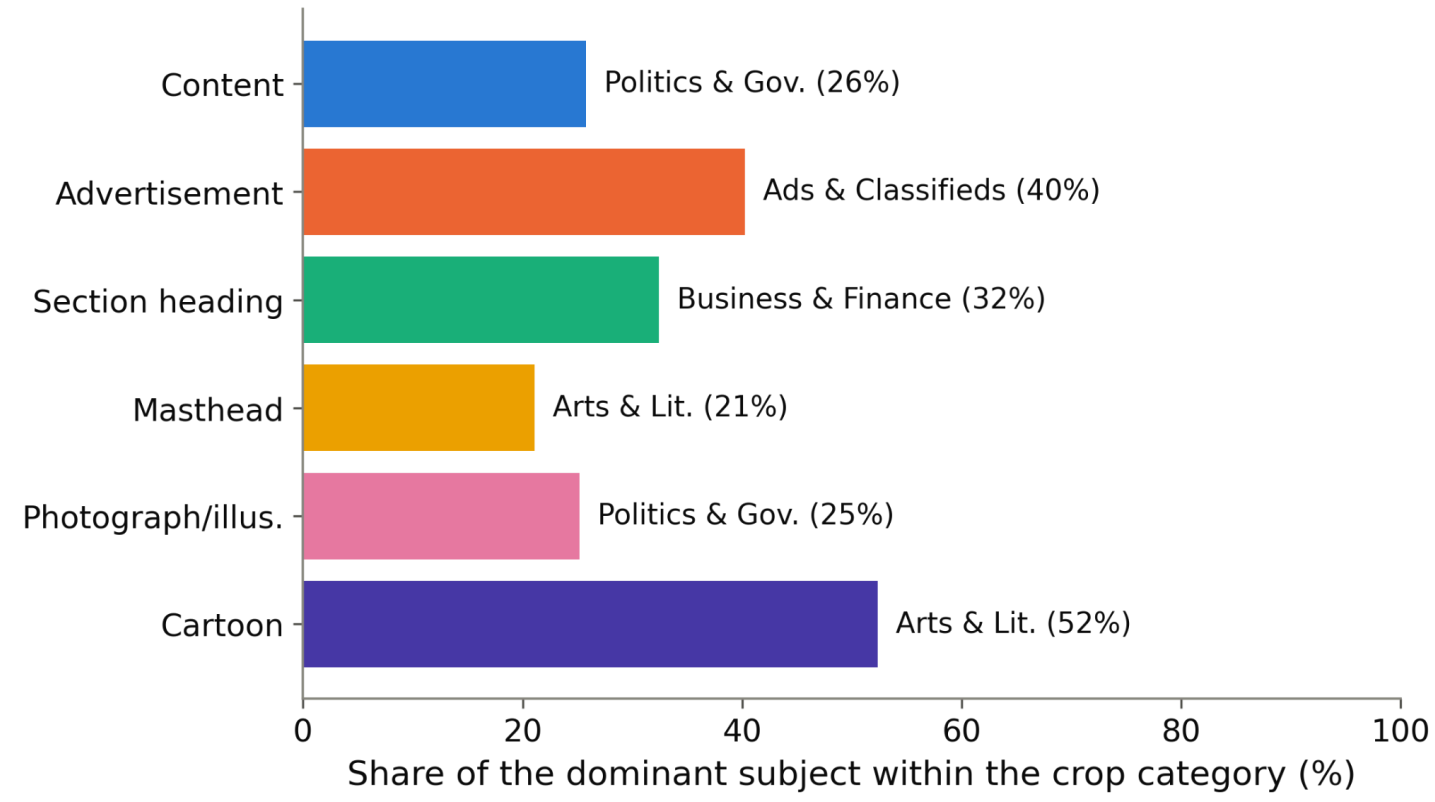

***Figure 16.*** *Dominant subject within each crop-type category (share of the leading subject, %).*

# 10 Crop-level Embeddings

## 10.1 Methodology

We hypothesize that pre-computed embeddings can improve the out-of-the-box accessibility of a dataset. As such, we chose to generate embeddings for both the text and the image of each crop as a way to enable dataset-wide vector search and, more generally, to facilitate downstream use of this dataset.

For image embeddings, we chose `facebook/dinov2-small` (Oquab et al., 2024) for its small size (22M parameters) and its relative genericity. We resized each crop to 448×448 pixels before inference.

For text embeddings, we chose `minishlab/potion-multilingual-128M` run through `Model2Vec` for its light footprint (a 128M-parameter static model) and its overall genericity.

## 10.2 Results

Both embedding sets are stored as 32-bit floating-point vectors, one per crop. The DINOv2-small produces 384-dimensional vectors and the potion-multilingual 256-dimensional vectors. As such, a single crop's embeddings occupy roughly 384 × 4 + 256 × 4 = 2,560 bytes. Across all 83.1 million crops, this is approximately 128 GB for the image embeddings (about 1.54 GB per million crops) and approximately 85 GB for the text embeddings, for a combined total of roughly 213 GB.

While this type of pre-computed output is likely to be helpful, it also has known limitations. The static text embedding model we used has no attention mechanism and therefore relies heavily on token distribution, reducing its usability in semantically ambiguous contexts. Both models are generic and may not support all use cases. As such, specialized applications will likely benefit from embeddings generated with domain or task-specific models.

# 11 Crop-level Detection of Chronicling America's Thesauri Terms

In collaboration with Harvard Law School Library's Public Data Project[10], we made use of thesauri on race, ethnicity, immigration, and citizenship (Gilmore, 2022), originally compiled by Library of Congress, National Endowment for the Humanities and state partners, and later preserved by the Public Data Project (Harvard Law School Library Innovation Lab, 2026). We ran detections for these terms against the VLM-extracted text of each crop.

The goal was to facilitate digital humanities research use cases, for example on contextualizing historical content and language. We surface these matches as a navigational and statistical aid rather than as an interpretation of the underlying text. These detections are both experimental and "naive"; they should not be used "as is" to infer the nature of any given piece of text.

[10] https://lil.law.harvard.edu/our-work/public-data-project/

# 12 Pipeline Footprint

## 12.1 Methodology

The pipeline was run locally on a GPU node we own (described in Appendix D). We designed the pipeline so it can be run on workstation-grade hardware, with throughput as the main trade-off. In any case, running this pipeline against Boston Public Library's collection represented a significant computational undertaking.

In some cases, we had to deploy optimization strategies to make the best of our hardware. Specifically, we identified a bottleneck early on when running inference with YOLO26 models through the `ultralytics` Python library. We noticed that, while the library allows for batched inference, image pre-processing appeared to happen sequentially rather than in parallel. We therefore re-implemented that part of the pipeline to pre-process images on our end, in parallel, before feeding them to the model as a pre-processed batch of matrices.

Retrieving scans and decoding them was also part of the challenge. To make this work, we processed the collection in batches of 200 issues and used a local disk cache with `python-diskcache` (Jenks, 2023), so that scan retrieval and decoding for one batch could overlap with the compute-heavy steps and would not have to be repeated.

## 12.2 Results

Processing time was dominated by the two OCR steps (Table 9). On a batch of 200 issues, dots.mocr OCR took 28.62 minutes on average and Tesseract OCR 16.48 minutes, together accounting for roughly half of the 86.69-minute average full batch. The remaining thirteen steps were comparatively inexpensive, most of them running in a few minutes or less.

While we deployed this pipeline on local hardware (Appendix D), we estimate the total cost of running it against this collection on rented hardware to be approximately ~$25,000, assuming a total runtime of ~1650 hours and the cost of a rented 8xL40S GPU node to be ~$15/hour on average (as estimated at the time of writing this technical report).

| **Pipeline step** | **Avg (min)** | **Min (min)** | **Max (min)** |
|---|---|---|---|
| 01 Scan retrieval & caching | 8.07 | 4.94 | 35.60 |
| 02 Crop detection (segmentation) | 8.14 | 6.18 | 22.74 |
| 03 OCR (Tesseract) | 16.48 | 12.28 | 22.79 |
| 04 OCR (dots.mocr VLM) | 28.62 | 22.46 | 35.90 |
| 05 Classification (text) | 1.01 | 0.79 | 1.30 |
| 06 Classification (image) | 4.48 | 1.80 | 5.75 |
| 07 Classification (final) | 0.09 | 0.06 | 0.27 |
| 08 Named entity recognition | 4.55 | 3.44 | 7.20 |
| 09 Subject detection | 5.04 | 4.18 | 6.19 |
| 10 Reading order | 1.13 | 0.90 | 1.56 |
| 11 Token counting | 1.28 | 0.72 | 2.42 |
| 12 Language detection | 0.90 | 0.68 | 1.21 |
| 13 Text analysis | 2.36 | 1.13 | 4.61 |
| 14 Chronicling America thesauri match | 2.17 | 1.03 | 3.19 |
| 15 Embeddings | 2.37 | 1.77 | 3.15 |
| **Full batch (200 issues)** | 86.69 | 68.20 | 111.49 |

***Table 9.*** *Average, minimum, and maximum processing time per pipeline step, per batch of 200 issues. Runs that crashed and/or ran faster because of pre-caching (e.g, after a crash) were not included. Durations for the "Full batch" row were measured directly from logs, not derived from per-step min/max/averages.*

## 13 Dataset Preparation and Post-Processing

The pipeline was designed to record data points into a SQLite database as it runs. These records were then processed into a dataset, exported as Parquet[11] shards of 250 scans each (one row per scan), compressed with `zstd` (Collet and Kucherawy, 2021) and written with a row-group size of 10. The field list is provided in Appendix E.

As we compiled the dataset, we applied some light post-processing to enhance usability:

- Minor text-processing was applied to the VLM OCR text, including: regular-expression-based dehyphenation, Markdown heading normalization, and token-loop removal. The latter, in which a word or character is repeated many times over, is a known failure mode of VLM OCR (Holtzman et al., 2020).

[11] https://parquet.apache.org

- We corrected likely errors in the locality metadata derived from the Library of Congress' API. For example, cases in which the city of Boston was associated with the state of New York.
- We applied filtering rules to language detection. We used the issue-level language code instead of our language detection for crops where the confidence score was below 0.50, or where the word count was below 30. In addition, we defaulted to issue-level language if said language was not supported by Lingua, which was the case for Yiddish.

The locality and issue-level language metadata that the last two rules rely on comes from the Library of Congress API (Library of Congress, 2026), which serves catalog records for `loc.gov` items as JSON.

These rules replaced the detected language code for 2,521,901 crops, or 3.04% of the crops that carry one. Replaced codes are stored without a confidence score so they remain distinguishable from direct detections. The data presented in this report reflects this post-processing.

## 14 Discussion and Future Directions

We see this work as a meaningful step towards unlocking high-quality data from historical newspapers, an important and abundant resource that remains underused in the age of computational access.

Our current set of models and methods was developed and tested against a subset of Boston Public Library's substantial newspaper collection. As a result, it is likely that our pipeline will struggle to process newspaper scans that are meaningfully different from what is present in this collection. This is a limitation we aim to address by continuously revising our models and methods, and by repeating this experiment on different collections with our library partners. Some of our methods have known, collection-independent limitations that we aim to soften over time. For example, our reading order detection mechanism (Section 7) makes assumptions about layout structure, and is therefore unlikely to be a good fit for newspaper scans that are less strictly column-based.

As a future direction, we also intend to further reduce the computational footprint of our pipeline. In particular, VLM-based OCR at this scale remains a significant bottleneck. We hypothesize that training a VLM specifically for performing OCR on historical newspaper crops could produce a model that is both smaller, more accurate and less sensitive to token looping issues than dots.mocr for this specific task.

We look forward to seeing what the community does with this data, these models, and this pipeline. We welcome suggestions and feedback, and hope to work together to further amplify the impact of this medium.

## Rights determination

We respect the intellectual property rights of authors, publishers, and other rights holders. We have taken deliberate steps to include only those issues for which there is no known copyright restriction. The source materials used in the preparation of this dataset were published before 1931, making them public domain in the United States. Issue-level metadata provided by Boston Public Library was used to make this assessment.

While this is relatively low risk, some materials in this dataset may be in the public domain in the United States but still subject to copyright or other rights protections in other jurisdictions. Additionally, the absence of an explicit copyright claim or rights status does not guarantee that a work is in the public domain, either in the U.S. or abroad. Information about the copyright status of individual issues is provided on a good-faith basis and reflects available data at the time of determination, but we cannot guarantee its completeness or accuracy.

Users of this dataset will be solely responsible for making independent legal assessments about how and where they use the materials. Some uses of materials may also be restricted by trademark, privacy, publicity rights, or other such rights or restrictions. It is the user's sole responsibility to consider the possibility that such rights or restrictions may be involved and to secure any needed permissions. If any rights holder believes that a work included in this release is misidentified or improperly included, we welcome contact and will promptly review any concerns. Our goal is to provide broad public access while maintaining respect for intellectual property rights and ensuring responsible data stewardship.

## Acknowledgements

The authors of this technical report would like to thank:

- The Digital & Online Services team at Boston Public Library for their expertise and continuous feedback, which shaped this work.
- Our team at the Institutional Data Initiative for their help with our initial rounds of annotations.
- Prof. Benjamin Charles Germain Lee and Dr. Molly Hardy for their advice and support.

This work was supported by unrestricted funding from Microsoft, OpenAI, Meta and Jane Street.

AI tools were used to assist in the preparation of this technical report.

## Disclaimers

### Harmful Language and Content in this Dataset

This dataset is a collection of historical works that reflect the language, culture, and perspectives of their time. Users should be aware that some materials may contain language or portrayals that are outdated, offensive, or harmful today, such as racism, sexism, colonial attitudes, and other forms of discrimination. Some content may include inaccurate information, providing insight into historical contexts that existed at the time of writing. The materials are maintained in their original form to retain contextual understanding and facilitate research efforts, but we encourage critical awareness and cultural sensitivity for the creators and/or subjects of the collection. These materials are offered as part of a historical perspective, but should not be considered a stand-alone research collection constructed to give a balanced perspective on any topic.

## Harmful Language in Bibliographic Description

Metadata for this collection may contain language that is overtly or implicitly harmful, outdated, or biased, or may by omission fail to represent important perspectives. Metadata may contain language created decades ago. It is common practice within the field of library science to reuse descriptions provided from the creator of the materials. While in some instances this allows communities and individuals to represent their materials in their own words, unexamined use of this practice may mean that racist or other offensive terminologies appear in our description. We also use national standardized terms in our work that can be outdated and harmful. Note that terminology in historical materials and in library descriptions does not always match the language we currently understand to be preferred by members of the communities depicted.

Furthermore, we acknowledge that the act of collecting materials is not always neutral, and the work of describing and classifying library materials is influenced by inherent personal, institutional, and societal biases. Outdated or offensive terminologies may be present in metadata such as subject headings, and harmful language or bias may be introduced by catalogers supplying titles and descriptions. In other cases, the source materials themselves present racist, offensive or otherwise harmful viewpoints in titles or descriptions that are routinely transcribed by catalogers.

**Note:** Some language in this statement was adopted from Harvard Library's statement on Harmful Language in Library collections[12].

## Generated and Experimental Content

This dataset contains generated and/or experimental content. While reasonable care was taken to ensure its quality, it is provided “as is,” without warranties of any kind. It may contain errors or inaccuracies; users should verify the data independently and apply their own judgment.

[12] https://library.harvard.edu/harmful-language-library-collections

# Appendices

## Appendix A — Outer scan margins

Section 3.2 reports that crop detections cover 89% of each scan on average. To determine how much of the remaining area is empty margin rather than missed content, we measured the distance between each edge of the scan and the outermost crop bounding box on a random sample of 100 scans (one scan drawn from each of 100 randomly selected metadata shards, seed 0). No sampled scan had to be discarded for missing dimensions or for having no crops. Bounding boxes are stored in source-scan pixels, so the left and top margins are the smallest `x0` and `y0` across all crops of the scan, and the right and bottom margins are the scan width and height minus the largest `x1` and `y1`.

| Margin | Mean (px) | Min (px) | Max (px) | Mean (% of scan) | Min (%) | Max (%) |
|---|---|---|---|---|---|---|
| Left | 228 | 0 | 690 | 3.06% | 0.00% | 7.37% |
| Top | 253 | 8 | 1,054 | 2.57% | 0.12% | 10.29% |
| Right | 264 | 7 | 550 | 3.63% | 0.10% | 9.96% |
| Bottom | 302 | 20 | 704 | 3.14% | 0.21% | 7.81% |

***Table A.1.*** *Outer margins left uncovered by crop detections, across 100 randomly sampled scans.*

The rectangle that encloses all crops of a scan covers 88.02% of the scan area on average, with a minimum of 75.84% and a maximum of 95.92%. That envelope is slightly smaller than the 89% average coverage reported in Section 3.2, which suggests the crops of a scan fill their own envelope almost completely. These results indicate that uncovered area therefore sits mainly in the outer margins, and the segmentation model leaves comparatively little content uncovered inside the printed area of the page.

## Appendix B — Crop-type auto-annotation prompt

The following prompt was used with `Qwen/Qwen3-VL-30B-A3B-Thinking-FP8` to auto-annotate the crop-type classification training set (see Section 6.1). The OCR text of the crop is appended after the Text: marker.

```
Carefully analyze the following newspaper cut and its text to determine its type.
Consider the content, structure, purpose, and context in which the content is
presented, including layout, formatting, and any visual elements.

The type can be one of the following:
- Content
- Section heading
- Advertisement
- Photograph or illustration
- Cartoon
- Masthead, nameplate or running head
- Empty

Classification guidelines:
- "Content": Includes both complete and incomplete articles, notices, schedules
(e.g., TV, Radio, train, ferries), stock listings, letters to the editor, games and
literary miscellany (poems, stories, essays, etc).
- "Section heading": Contains only the title of a section with no additional text.
Generally short. Article titles should be labelled as "Content".
- "Advertisement": Includes any content promoting a product, service, company or
event. Electoral advertisements should be considered "Advertisements" unless
they're part of an official notice.
- "Photograph or illustration": May include a title and/or legend. If part of an
advertisement, classify as "Advertisement." If the cut contains only an image with
no text, classify as "Photograph or illustration" unless it's part of an
advertisement. A cut containing only text cannot be "Photograph or illustration".
- "Masthead, nameplate or running head": A cut that displays the newspaper's title
and/or logo, and may include other a page number, and/or date. Not to be confused
with "Section heading".
- "Empty": Empty cut. Contains no text and no clear visual element.

When classifying, please consider the following:
- If a cut contains multiple types of content, classify it based on the primary or
dominant content. If two types of content are equally prominent, prioritize the one
that is most relevant to the overall purpose of the cut.
- If you're unsure about the classification, consider the context in which the
content is presented and the purpose it serves.

Return the most appropriate type for this cut and nothing else.

Text:
```

***Figure B.1.*** *Prompt used to auto-annotate the crop-type classification training set (Section 6)*

## Appendix C — Crop-type classifier confidence

Section 6.2 reports the overall confidence of each crop-type classifier (text and image). This appendix breaks those figures down by category.

Confidence is highest on the two categories that carry almost all of the volume, “Advertisement” and “Content”, and on the two structural categories, ‘Section heading” and “Masthead, nameplate or running head”. It drops sharply on “Photograph or illustration” and “Cartoon”, the two categories where Table 4 also reports the weakest precision and recall. Both classifiers therefore signal their own uncertainty on the same categories.

The text classifier does not model “Empty” as a learned label: as noted in Section 6.1, every crop with no OCR-able text is assigned “Empty” by rule. Its 1.000 ± 0.000 is therefore an artifact of that rule and not a measurement. The image classifier does predict “Empty” as a learned label, and its 0.679 mean is the lowest of the seven, which is consistent with the low precision reported for that label in Table 4.

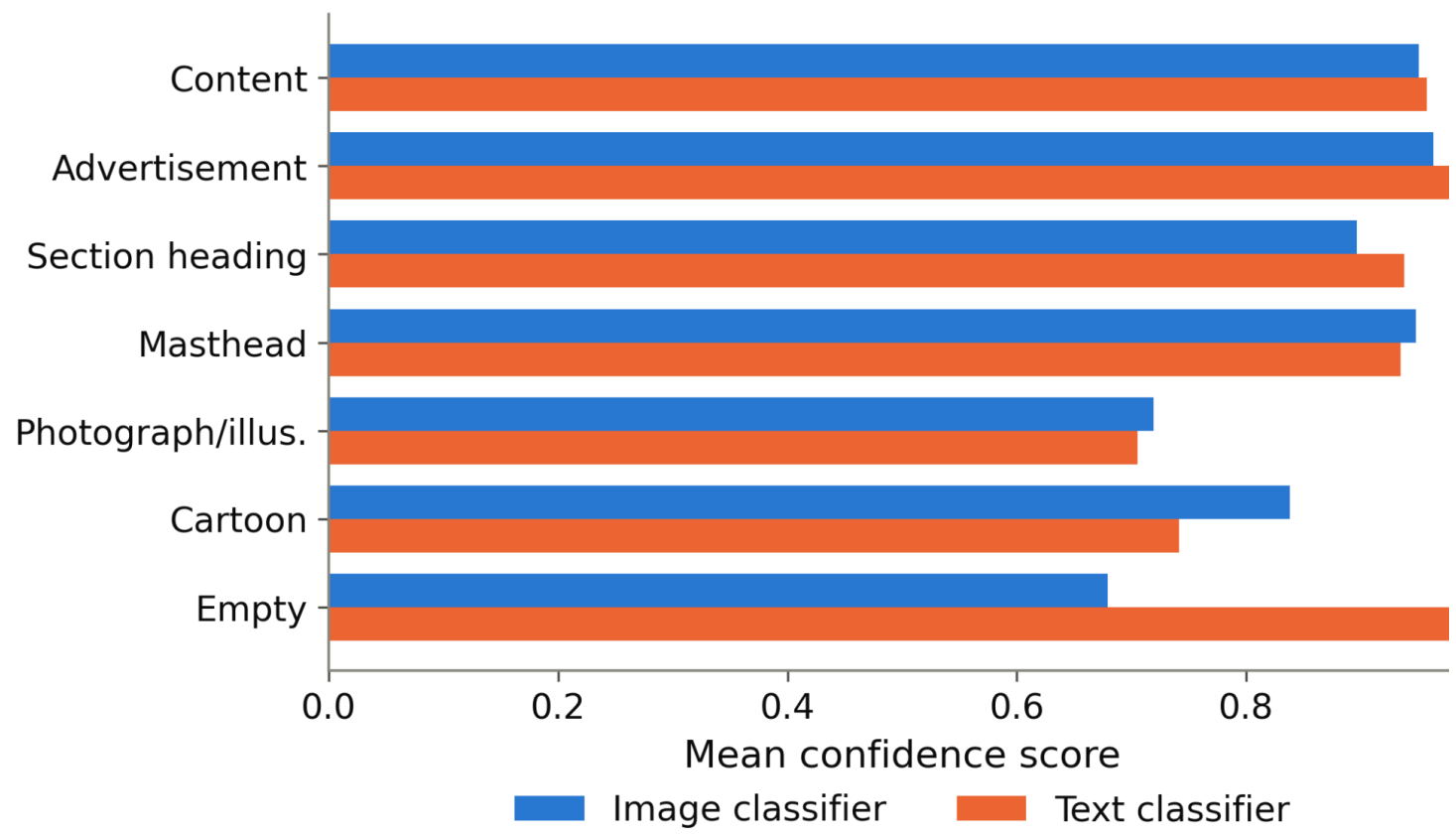

***Figure C.1.*** *Mean confidence score of the image and text crop-type classifiers, per predicted category.*

| **Category** | **Image mean ± SD** | **Image crops** | **Text mean ± SD** | **Text crops** |
|---|---|---|---|---|
| Content | 0.950 ± 0.107 | 26,973,099 | 0.957 ± 0.102 | 28,898,044 |
| Advertisement | 0.963 ± 0.089 | 49,025,100 | 0.978 ± 0.076 | 48,095,943 |
| Section heading | 0.896 ± 0.131 | 5,045,743 | 0.938 ± 0.125 | 4,091,158 |
| Masthead, nameplate or running head | 0.948 ± 0.130 | 1,414,387 | 0.935 ± 0.138 | 1,567,417 |
| Photograph or illustration | 0.719 ± 0.141 | 423,011 | 0.705 ± 0.184 | 227,434 |
| Cartoon | 0.838 ± 0.202 | 229,046 | 0.742 ± 0.197 | 105,970 |
| Empty | 0.679 ± 0.173 | 36,655 | 1.000 ± 0.000 | 161,075 |
| All categories (weighted) | 0.953 | 83,147,041 | 0.967 | 83,147,041 |

***Table C.1.*** *Mean confidence score and standard deviation of each crop-type classifier, per predicted category, across all classified crops.*

## Appendix D — Compute environment

The pipeline was run on a single GPU node owned by the Institutional Data Initiative, with the following specification:

- 8 × NVIDIA L40S GPUs
- 256 CPU cores
- 768 GB of RAM

## Appendix E — Dataset field list

Each row in the dataset represents a single newspaper scan (page). Most crop-level fields are list columns, with one entry per crop in the same order as `crop_bbox_gen`, which is sorted by reading order. NER, classification, subject, and language fields are lists of lists (one inner list per crop).

| Suffix | Meaning |
|---|---|
| `_src` | *"From source"*. This field's data comes from information we gathered from the collection itself. |
| `_ext` | *"External"*. This field's data was pulled from an external source via a records matching mechanism. |
| `_gen` | *"Generated"*. This field's data was generated as part of our analysis / processing pipeline. |
| `_exp` | *"Experimental"*. Similar to _gen, but was generated through more experimental or exploratory means. |

***Table E.1.*** *Suffix convention for column provenance.*

| Field | Type | Role | Section |
|---|---|---|---|
| `scan_image` | `bytes` | The source scan, stored as embedded WEBP bytes at quality 70. | 1 |
| `corpus` | `string` | Corpus identifier | 13 |
| `issue_id_src` | `string` | Issue identifier | 2 |
| `newspaper_id_src` | `string` | Newspaper (title) identifier | 2 |
| `newspaper_id_type_gen` | `string` | Type of the newspaper identifier. | 2 |
| `page_number_gen` | `int` | Page number within the issue. Inferred from filenames within source archive | 2 |
| `scan_filename_src` | `string` | Source scan filename | 2 |
| `scan_width_src` | `int` | Scan width in pixels | 3 |
| `scan_height_src` | `int` | Scan height in pixels | 3 |

| year_ext | int | Publication year | 2 |
|---|---|---|---|
| month_ext | int | Publication month | 2 |
| day_ext | int | Publication day | 2 |
| edition_ext | int | Edition | 2 |
| metadata_source_gen | string | Source of the issue-level metadata | 5, 13 |
| city_gen | string | Locality city (post-processed, sourced from metadata_source_gen) | 13 |
| state_gen | string | Locality state (post-processed, sourced from metadata_source_gen) | 13 |
| country_gen | string | Locality country (post-processed, sourced from metadata_source_gen) | 13 |
| language_ext | string | Issue-level language code (sourced from metadata_source_gen) | 5 |
| crop_bbox_gen | list[list[float]] | Crop bounding boxes | 3 |
| crop_bbox_conf_gen | list[float] | Crop detection confidence scores | 3 |
| crop_vlm_ocr_gen | list[string] | dots.mocr OCR text (post-processed) | 4 |
| crop_tesseract_ocr_gen | list[struct] | OCR text of each crop from Tesseract 5, as {text, metadata}. text is plain text. metadata is a list holding one record per recognized word, each with text (the surface form), conf (the Tesseract confidence score), and bbox_xyxy (the word box as [x_min, y_min, x_max, y_max], in pixel coordinates relative to the crop, rescaled back to the crop's full resolution). Useful for building OCR outputs compatible with existing library discovery systems. | 4 |
| crop_vlm_ocr_token_count_gen | list[int] | dots.mocr o200k_base token count | 4 |
| crop_tesseract_ocr_token_count_gen | list[int] | Tesseract o200k_base token count | 4 |
| crop_text_analysis_gen | list[list[string]] | Per-crop text metrics. Holds tokenizability_score, char_count, word_count, word_count_unique, word_type_token_ratio, sentence_count, and sentence_count_unique, each prefixed tesseract_ and vlm_, plus vlm_has_table and vlm_has_markdown. | 5 |
| `crop_classification_gen` | `list[string]` | Final crop-type category | 6 |
| `crop_classification_image_only_gen` | `list[string]` | Image-classifier category | 6 |
| `crop_classification_image_only_conf_gen` | `list[float]` | Image-classifier confidence | 6 |
| `crop_classification_text_only_gen` | `list[string]` | Text-classifier category | 6 |
| `crop_classification_text_only_conf_gen` | `list[float]` | Text-classifier confidence | 6 |
| `crop_language_gen` | `list[string]` | Detected language (post-processed, issue-level language used if confidence is low) | 5 |
| `crop_language_conf_gen` | `list[float]` | Detected language confidence (post-processed, NULL if issue-level language was used) | 5 |
| `crop_ner_per_gen` | `list[list[string]]` | Person entities | 8 |
| `crop_ner_per_conf_gen` | `list[list[float]]` | Person entity confidence | 8 |
| `crop_ner_loc_gen` | `list[list[string]]` | Location entities | 8 |
| `crop_ner_loc_conf_gen` | `list[list[float]]` | Location entity confidence | 8 |
| `crop_ner_org_gen` | `list[list[string]]` | Organization entities | 8 |
| `crop_ner_org_conf_gen` | `list[list[float]]` | Organization entity confidence | 8 |
| `crop_subject_gen` | `list[list[string]]` | Subject ranking | 9 |
| `crop_subject_conf_gen` | `list[list[float]]` | Subject confidence | 9 |
| `crop_chronam_thesauri_matches_exp` | `list[struct]` | Keyword matches from the Race, Ethnicity, Citizenship and Immigration Keyword Thesauri for Chronicling America. Holds matches (a list of `{category, terms}`, terms being `{term, count}` pairs), `match_count`, and `term_count`, each prefixed `tesseract_` and `vlm_`. Naive match: may be used for downstream research, but not "as is". (See section 11) | 11 |
| `crop_text_embeddings` | `list[list[float]]` | Text embeddings (potion-multilingual, 256-d) | 10 |
| `crop_image_embeddings` | `list[list[float]]` | Image embeddings (DINOv2-small, 384-d) | 1 |

***Table E.2.*** Dataset fields with their type, grouped by role, and the report section each relates to.